%% file: 00_InvestigationChessPlayerAttentionPrediction.tex
\documentclass[letterpaper, 10pt]{article}
\usepackage[letterpaper,
             left=3cm,
             right=3cm,
             top=2.5cm,
             bottom=2.5cm
             ]{geometry}

\usepackage{soul}
\usepackage[square]{natbib}

\usepackage[utf8]{inputenc}
\usepackage{booktabs} %
\usepackage{tikz}
\def\Checkmark{\tikz\fill[scale=0.4](0,.35) -- (.25,0) -- (1,.7) -- (.25,.15) -- cycle;}

\usepackage{tabularx}
\usepackage{float}
\usepackage{makecell}
\usepackage{enumitem}
\usepackage{rotating}

\newcolumntype{Y}{>{\centering\arraybackslash}X}

\usepackage[ruled]{algorithm2e} %

\SetAlFnt{\small}
\SetAlCapFnt{\small}
\SetAlCapNameFnt{\small}
\SetAlCapHSkip{0pt}
\IncMargin{-\parindent} 

\newcommand{\etdatasetinit}{\textit{ETDS}}
\newcommand{\augetdatasetinit}{\textit{ETDSa}}
\newcommand{\gamesdatasetinit}{\textit{GDS}}
\newcommand{\etdataset}{\etdatasetinit{}}
\newcommand{\augetdataset}{\augetdatasetinit{}}
\newcommand{\gamesdataset}{\gamesdatasetinit{}}

\newcommand{\beforemainsection}{}

\usepackage[colorlinks=true,citecolor=blue,urlcolor=blue,linkcolor=blue]{hyperref}

\newcommand{\added}[1]{#1} %

\makeatletter
\newif\if@anonymize

\@anonymizefalse  %

\if@anonymize
\usepackage{tikz}
\usetikzlibrary{calc}
\usepackage{censor,caption}
\newcommand{\anonymizeemail}[1]{none@anonymize.invalid}
\newcommand{\anonymizebib}[1]{XXXXXXXXXX}
\newcommand{\anonymizeurl}[1]{http://www.xxx.xxx/}
\else

  \newcommand{\anonymizeemail}[1]{#1}
  \newcommand{\anonymizebib}[1]{#1}
  \newcommand{\anonymizeurl}[1]{#1}
\fi
\makeatother

\usepackage{color}
\usepackage{xcolor}

\renewcommand\cite[1]{\citep{#1}}

\hypersetup{
  pdftitle={Deep learning investigation for chess player attention prediction using eye-tracking and game data},
  pdfauthor={Justin Le Louedec, Thomas Guntz, James L. Crowley, Dominique Vaufreydaz},
  pdfkeywords={Deep neural network, Computer vision, Visual attention, Chess},
}

\title{Deep learning investigation for chess player attention prediction using eye-tracking and game data}

\author{Justin Le Louedec, Thomas Guntz, James L. Crowley, Dominique Vaufreydaz}

\date{
\small{Univ. Grenoble Alpes, CNRS, Inria, Grenoble INP, LIG, 38000 Grenoble, France}\\
\vspace{0.5em}\textit{Author Version}
}

\begin{document}

\maketitle

\begin{abstract}

This article reports on an investigation of the use of convolutional neural networks to predict the visual attention  of chess players.  
The visual attention model described in this article has been created to generate saliency maps that capture hierarchical and
spatial features of chessboard, in order to predict the probability fixation for individual pixels
Using a skip-layer architecture of an autoencoder, with a unified decoder, we are able to use multiscale features to predict saliency of part of the board at different scales, showing multiple relations between pieces.
We have used  scan path and fixation data from players engaged in solving chess problems, 
to compute 6600 saliency maps associated to the corresponding chess piece configurations. 
This corpus  is completed with synthetically generated data from actual games gathered from an online chess platform. Experiments realized using  both scan-paths from chess players and the CAT2000 saliency dataset of natural images, highlights several results. Deep features, pretrained on natural images, were found to be  helpful in training visual attention prediction for chess. The proposed neural network architecture is able to generate meaningful saliency maps on unseen chess configurations with good scores on standard metrics. 
This work provides a baseline for future work on visual attention prediction in similar contexts.

\end{abstract}

\maketitle
\begin{center}
\end{center}

\input{01_Introduction.tex}

\input{02_Background.tex}
\input{03a_OurApproach.tex}

\input{03b_Training_data.tex}

\input{05_Results.tex}

\input{06_Discussion.tex}

\input{07_Conclusion.tex}
\input{08_ACKS.tex}

\bibliographystyle{ACM-Reference-Format} %
\bibliography{Bibliography}

\end{document}

%% file: 01_Introduction.tex
\section{Introduction}

Humans possess a remarkable ability to discriminate and select relevant information from an environment without attending to the entire scene. This ability for selective attention enables very fast interpretation and rapid interaction. Models for this process are of growing interest for both computer vision and neurosciences. Modelling visual attention not only provides an understanding of how the human visual system processes information but also has practical applications for constructing artificial systems for object recognition,  event detection   \cite{saliencydetection,saliencydetectionvideo,Buso2015}, action recognition \cite{DBLP:journals/corr/SafaeiF17,phdthesis}, object segmentation \cite{10.1007/978-3-319-01796-9_31,7984578}, as well as computer graphics \cite{3Dshapevisualattention,phdthesis2}.

\begin{figure*}[ht]
    \centering
    \includegraphics[width=0.32\linewidth]{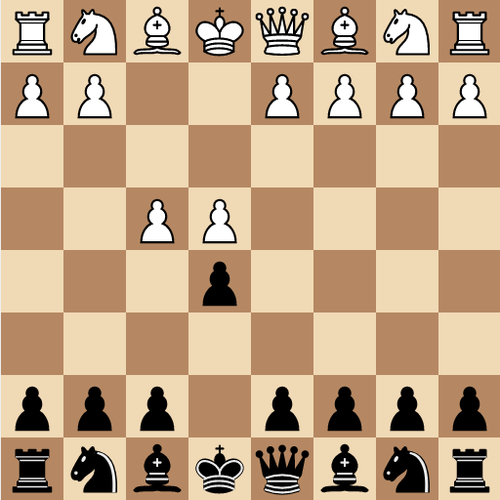}\hfill
    \includegraphics[width=0.32\linewidth]{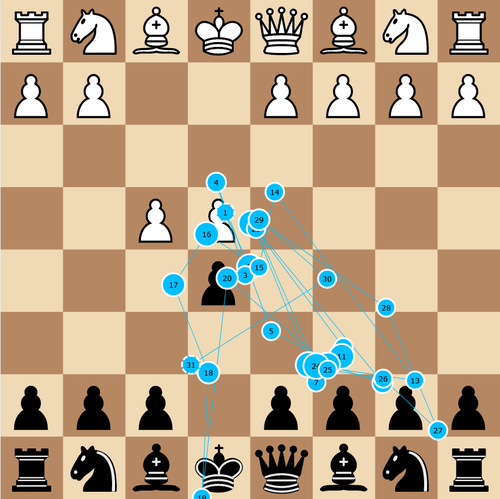}\hfill
    \includegraphics[width=0.32\linewidth]{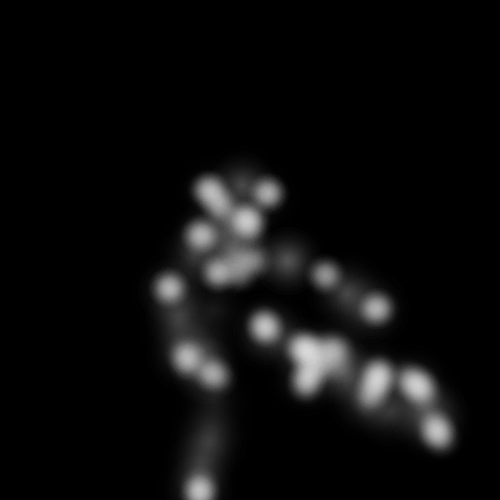}
    \vspace{-1em}
    \caption{Saliency map from a chess game. At left, input chess board image. At centre, eye tracking results (points represent eye fixations, lines are scan path between fixations). At right, saliency map computed from eye tracking. For each pixel, a probability between 0 (black) and 1 (white) is computed.}
    \label{fig:saliency_chess}
    \vspace{-1em}
\end{figure*}

Most work on Visual attention prediction has attempted to quantitatively or qualitatively predict gaze location.
These locations are often displayed in the form of saliency maps, i.e. a map that displays the probability that each pixel may be fixated (receive visual attention).
For example, the right side of figure \ref{fig:saliency_chess} shows a saliency map computed from eye-tracking data.
In this example, one can see that whiter parts of the saliency maps are more frequently fixated than others.
Aggregated eye-tracking information from many users can be used to estimate the probability that a pixel will be fixated. However, accurate prediction requires an extensive amounts of eye-tracking data.

Current techniques for saliency or visual attention prediction fall into two main categories: Bottom-Up and top-Down approaches.
Bottom-Up approaches are common in computer vision  where, local image appearance is assumed to stimulate fixation, and the goal is to provide  a probability that each pixel will be fixated without prior knowledge of the task or task context.
Top down approaches are generally task-driven and require an understanding of the image, scene, context, or task.
Visual attention in Chess can be seen as a combination of these two approaches as salient areas may be defined by both visual pattern recognition (bottom-up analysis) and an understanding of the game (top-down prediction) \cite{Charness2001,Perceptioninchess,doi:10.1167/17.3.4}. %
This suggests the extraction of local features and interpretation from images  that are very different from the natural images most commonly used in visual attention prediction.
In particular, the saliency of chess pieces depends on the current game configuration,  expressed as configurations of pieces (Figure \ref{fig:saliency_chess}).

In this paper, we address the problem of visual attention prediction for chess players using a convolutional neural network (CNN). While CNN's have been used for attention prediction in non-natural image such as comics~\cite{Bannier:2018}, we believe that this is first investigation for the use in predicting attention in chess. 
Our investigation  builds on previous works on visual attention \protect\cite{7410395,DBLP:journals/corr/KruthiventiAB15,DBLP:journals/corr/WangS17b} and Chess expertise \protect\cite{Charness2001,Perceptioninchess,doi:10.1167/17.3.4}, to create a network architecture capable of predicting visual attention of chess players.
We examine several training strategies, and evaluate interest of different training corpora including a state-of-the-art saliency corpus, a bottom-up corpus from actual eye-tracking data from chess players and a top-down corpus from online games built in the context of the project.

The article is organized as follow. Section \ref{chapter:background} provides background  about the problem of salience prediction. Section \ref{chapter:approach} describes our approach, choices and our model.
In section \ref{chapter:learning}, we describe the importance of data and the creation of our datasets. In section \ref{chapter:results}, we detail performance results of different training strategies using attention prediction metrics and qualitative results from different chess configurations. In a final section, we  discuss the results and limitations.

%% file: 02_Background.tex
\beforemainsection{}
\section{Visual attention in Chess} \label{chapter:background}

Visual attention has evolved to enable rapid analysis of interesting and relevant regions in visual scene. This frees resources so that higher-level cognitive processes may  perform more complex tasks such as scene understanding or decision making from limited data. 

In their paper, \citet{posner1990} argue that visual attention is not the function of a single area of the brain but rather the result of a network of different components. He explains that visual recognition depends on successions of neural processes to extract information visual stimuli. 

Modern Eye tracking systems make it possible to record the locations in a visual scene that receive fixation as well as the path between  fixations  (scan path). Fixations are characterized by the time spent on different parts of the scene and  can be used as indicators on most viewed and fixated areas on images or scenes. They are often transformed to continuous saliency maps by convolution with a  Gaussian function where intensity indicates  duration of fixation. 

As said, figure \ref{fig:saliency_chess} shows an example of eye tracking data with a chess configuration in the background. On the center image, points indicate fixations and the lines between points are the scan path of gaze between fixations. This data was acquired for an experimentation on observing and interpreting subjects in problem solving tasks (chess in particular) through multimodal observations (eye tracking, posture, facial emotion, etc.) \cite{thomasguntz}.

\begin{figure*}
\begin{tabularx}{\linewidth}{@{}YY@{}}
\includegraphics[width=0.3\columnwidth]{./images/image.png}
\caption{Example of eye tracking data for an user on a chess configuration}\label{fig:exampleeye}
    &
    \includegraphics[width=0.45\columnwidth]{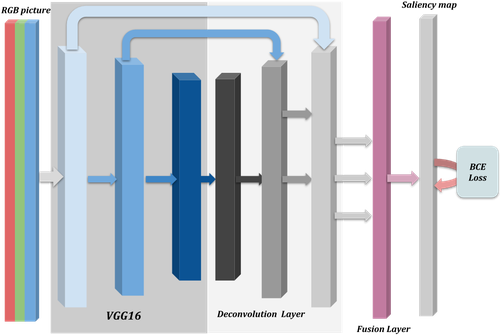}
\caption{Architecture of our Deep Neural Network}\label{fig:model}
\end{tabularx}
\end{figure*}

Eye tracking maps, captured from human subjects looking at  scenes with a small number of salient objects, show that humans tend to focus on a dominant object and ignore  background information.
Such an approach is not appropriate for Chess as the salient objects are  pieces, while their saliency is conditioned by tactical importance of their role in a configuration (Figure \ref{fig:saliency_chess}).
In addition, the saliency of individual chess pieces depends on the chess expertise of the player \cite{Perceptioninchess}.

Chess expertise and expertise in general are frequently studied in psychology.
\citet{Charness2001} discovered that experts fixate more often on empty squares than  intermediate players.
Experts also display a much larger number of fixations on relevant pieces than intermediate. It was argued that experts were encoding configurations rather than pieces and produced patterns of saccades (quick movement of eyes between two or more phases of fixation).
In their article, \citet{Perceptioninchess} demonstrated that expert players use larger visual spans while processing structured chess positions.
They also documented that  expert players would make fewer fixations and focus on relations between pieces rather than fixate a large number of individual pieces. 

In a more recent paper, \citet{doi:10.1167/17.3.4} proposed that chess expertise is mainly related to the detection of complex patterns and configurations. When improving their skills, players become capable of detecting increasingly complex patterns. This may explain why experts do not need to fixate pieces individually, and instead detect interesting patterns by looking at their center, using peripheral vision to detect the configuration. This specificity of visual attention of player depending on their skills is a difficulty to address in this research.

%% file: 03a_OurApproach.tex
\beforemainsection{}
\section{Deep Neural Network overview}\label{chapter:approach}
\subsection{Architecture Overview}

In the last few years, many deep learning architectures have been proposed for prediction of visual attention.
Deep learning solutions generally achieve better performance than traditional machine learning techniques.
For example,  \citet{7410395} propose a fine-tuning method   with objective functions and integrating information from the image at different scales, achieving top scores for standard metrics such as AUC-Judd or Linear correlation (respectively 0.87 and 0.74, see section \ref{sec:metrics} for metric descriptions).

Other models, such as Deepfix \cite{DBLP:journals/corr/KruthiventiAB15} or Deep visual attention \cite{DBLP:journals/corr/WangS17b}, have attempted to model visual attention using convolutional neural networks.
Deepfix focused on location dependent patterns using "Location biased convolution layer", concatenating Gaussian blobs to the inputs of convolutional layers.
Our work is inspired by 
Deep visual attention, which used an autoencoder network to generate saliency maps from images using skip layers and a split decoder.
Both studies provide good performances on free-view scenes of natural images on the MIT300 dataset \cite{Judd_2012} and made use of multi-scale features.

A particular strength of Convolutional Neural Networks (CNN) is to hierarchically capture  features  such as corners, edges and shared common patterns from raw images and to integrate such information to recognize more complex patterns such as faces, eyes, or the crown of a king.
This integration of  high and low level features is useful to capture characteristics of a piece.

The interaction of pieces at different positions on the board determines if an area is salient or not.
We base our approach on the work of \protect\citet{DBLP:journals/corr/WangS17b}, a multi-scale deep neural network encoding features from images to decode them as saliency maps.
The model is depicted in Figure \ref{fig:model}. As Wang and al. model, it is composed of an auto-encoder with skip layer connections. The encoder part of the model is a VGG16 \protect\cite{DBLP:journals/corr/SimonyanZ14a} pretrained on ImageNet \protect\cite{imagenet_cvpr09}, while the decoder part is a succession of deconvolution layers with RELUs, in charge of upsampling feature maps at different dimension/scale levels. 

There is a need to express the similarity in the relation between two pieces being close to each other or far away across the board.
This means that feature maps at the same dimension must be deconvolved the same way by the decoding part of the network.
To do that, we chose to combine the different deconvolution layers of Wang and al. network into only one decoding network. Deconvolution functions in deep learning are in fact transposed convolutions or fractionally strided convolutions which are mislabeled as deconvolution. In fact, these are simply spatial convolutions using a specific padding to up-sample in a learnable way the feature map. The use of deconvolutions rather than fixed kernel upsampling functions comes from learning to discriminate salient or not salient features. Doing so, the network becomes capable of extracting features from an image and deciding which ones are related to salient areas of an image. As an added benefit, this approach provides a higher computation efficiency by progressively reducing the feature space dimensions.

The decoding part of the deep network is a succession of deconvolution step on feature maps of different dimensions to up-sample the signal to the size of the input image. Three feature maps are taken at various stages of the encoding network and are decoded using the decoding network, giving 3 corresponding saliency maps. Those saliency maps are then combined and passed through a convolutional layer called a "fusion layer", which is reduced to form a final saliency map. This architecture makes it possible to learn the transformation of feature maps to salient areas and  to combine different scale saliency maps into a single global map describing the input image as displayed in figures \ref{fig:modelcomp} to \ref{fig:choice}.

\subsection{Training loss}
To train the network, one needs to define a loss function between two saliency maps.
The proposed model is close to a standard auto-encoder architecture which basically learns to encode some data and to decode it identically to the input. A direct approach to compare the output to the input would be to compute a distance metric that could be minimized  during the training process. 
The basic, simple and intuitive L1 loss ($|gt - f(x)|$ with $gt$ as the ground-truth and $f(x)$ the output of the auto-encoder) provides a possible solution. It would also be possible to use the L2 distance or Mean Square Error (MSE) expressed as $(gt- f(x))^2/N$ with $N$ being the batch size.

In our case, the network does not learn  to generate a copy of the input in the form of a chessboard with pieces on it, but the visual attention prediction from this chessboard. The visual prediction is a probability distribution ranges from  0 to 1 (0 not salient at all and 1 very salient).
The \textit{Kullback-Leibler} (KL) divergence \cite{10.2307/2236703} expresses the degree of divergence of two probability distributions. Minimizing the KL divergence is equivalent to reducing the distance between them. %
\added{However, the \textit{Kullback-Leibler} divergence is not useful as a metric, as it does not respect the triangle inequality: it is not symmetrical, it does not have an upper bound, and often $KL(P|Q) \neq KL(Q|P)$ as discussed in} \cite{LeMeur2013,metricsreview}.

\added{Binary Cross Entropy ($-gt.log(f(x))-(1-gt).log(1-f(x))$) is a good candidate in the case of saliency map generation.
It can be seen as relatively close to the \textit{Kullback-Leibler} divergence.
Indeed, cross entropy is often used to measure the dissimilarity between two distributions.
Minimizing cross entropy makes is possible to generate data as close as possible to the ground-truth, while avoiding \textit{Kullback-Leibler} divergence's problems.
Thus, Binary cross entropy is  used as a loss function in the experiments described below} \ref{chapter:results}.

%% file: 03b_Training_data.tex
\section{Training corpora}  \label{chapter:learning}

\subsection{Bottom-up datasets from eye tracking data (\etdataset~and \augetdataset)}
\subsubsection{Initial dataset}

The initial dataset comes from a multimodal study with people solving chess problems \cite{thomasguntz}. For these experiments, they recorded the eye-gaze of 30 players  (10-51 years, 2 female, age: $M=30.0; SD=12.9$) from various levels (from beginners to experts, with a majority of intermediate players), trying to solve 13 chess tasks of increasing difficulty in limited time:
\begin{itemize}[leftmargin=*]
    \item 2 \textit{openings} tasks: \textit{King\'s Gambit} and a variation of the \textit{Caro-Kann} defense.
    \item 11 \textit{N-Check-Mate} tasks (putting the enemy king in a position of checkmate in $N$ moves, $N$ ranges from 1 to 6) tasks. 
\end{itemize}

\added{
In this corpus, 9 players were active expert players with~\textit{Elo} ratings\footnote{~The \textit{Elo} system is a method to calculate rating for players based on tournament performance. Ratings vary between 0 and approximately 2850. \url{https://en.wikipedia.org/wiki/Elo_rating_system} (last seen 09/2018)}
ranged from 1759 to 2150 ($M=1937.3$; $SD=123.2$). %
For the intermediate players, the \textit{Elo} ratings ranged from 1100 to 1513 ($M = 1380.3$; $SD = 100.5$).
6 players were casual players who were not currently playing in club.
The average recording time per participant is 13:35 minutes ($MIN=4$:$54$, $MAX=23$:$54$, $SD=5$:$02$).
}

The saliency maps of the openings are not very informative.
Common openings are repetitive and well known by players. They tend to follow more their intuition at this point of the game rather than reasoning.
Thus, eye tracking data does not provide satisfactory information for our prediction task and we decided to focus our study on the 11 \textit{N-Check-Mate} tasks that require more cognitive processes to be solved.

Visual attention differs among chess players and depends on their level of expertise \cite{Charness2001}. 
These differences are reflected in saliency maps.
The primary experiments using these data exhibited that the saliency maps are too sparse to let a neural network converge. 
As it is commonly done in saliency map prediction from natural images \cite{Bannier:2018}, 
we decided to group each task into one saliency map, averaging all saliency maps from all players computed from eye tracking data. 
This computation provides a mean representation closer to most players' visual attention.
Doing so, the current corpus ends-up with 11 configurations and the same amount of saliency maps.
With such limited data, we needed ways to augment it while keeping data interesting and relevant.
This corpus will be referenced throughout the rest of the article as the \etdataset{} dataset. 

\subsubsection{Data augmentation}

Data augmentation is about creating variations of the dataset, to multiply the examples the network will learn on and help it generalizing its knowledge to more and more complex examples.
Usually, this includes many transformations such as rotating, translating, flipping, scaling or cropping a part of the image, or even adding some noise to the input in some adversarial training paradigms.
Several augmentation techniques are relevant in our context:
\begin{itemize}[leftmargin=*]
    \item The first augmentation is to cut the board into smaller pieces, similarly for the corresponding saliency map. So we let the network focus on specific parts of the board, \added{learning features at different scales}.
    We used a sliding window of 3x3, 4x4 and 5x5 squares on the chess board, creating various sub-views of the configuration. This augmentation multiplies the amount of data by 150. The goal here is to make the network learn smaller and more precise features at the scale of few cells. %
    \item The second data augmentation consists in inverting the color of each piece (black to white and \textit{vice versa}). Because nearly every configuration is possible while playing white and black, inverting colors adds diversity to the configurations the network encounter while training. %
    The amount of training data is multiplied by 2.
    \item The last augmentation changes the orientation of the board to the point of view of the other player. The underlying hypothesis is that, at any time, a chess player exploring his adversary possibilities looks at the board the same way the initial player does. Again, this augmentation increases the available training data by a 2 factor. 
\end{itemize}

While remaining small in regards of usual state-of-the-art corpora, the final \augetdataset{} (for augmented \etdataset{}) contains 6600 examples.

\beforemainsection{}
\subsection{Top-Down dataset generated from games~(\gamesdataset{})}\label{sec:Top-Down}

As stated in previous section, Top-Down approaches in visual attention prediction focus on relevant information in an image depending on the task to accomplish.
In chess, the task is well defined as it is to move, for a specific configuration, the piece that is most satisfying a condition which could be to win the game, to put the king in check, etc.
Because we are working with visual attention and saliency maps, we do not want to train a network to play chess (unlike Alphazero \cite{DBLP:journals/corr/abs-1712-01815}) but to generate saliency maps depending on the pieces chess players consider to play at each round, helping the network to learn relevant pieces for a given configuration.

From the beginning of the project, we used Lichess\footnote{\url{https://en.lichess.org/} (last seen 12/2018).}, an online and offline chess platform.
It allows to create personalized games and to play against other players or a computer. With thousands of registered players and millions of games played every month, this is a gold mine to get training data. They grant access to a monthly database in the PGN format (Portable Game Notation)\footnote{\url{https://en.wikipedia.org/wiki/Portable_Game_Notation} (last seen (12/2018).}. This database contains all games played during the past month (more than 22 Million for March 2018).
Because we are working on visual attention with images, we had to transform these PGN games into images. We developed a tool which reads games and, for each turn, creates the image board as seen from the white and black perspectives. 

At this point, we want to generate saliency maps of chess players' visual attention. We need to find a way to generate relevant and accurate saliency maps. We based the processing on three basic hypotheses. The first one being that a chess player is at least looking at two things: the cell where lies the piece he intends to move and its destination. As we know from the PGN data which piece is going to move, we highlighted in the saliency maps these two areas on the board. 
The second hypothesis is that chess players are quickly following the path taken by the moved piece to check collision and accuracy of its destination. Even if this is not as salient as starting and ending fixations, it could be considered as a succession of eye saccades.
We model it in the saliency map as a darker shade of gray showing its smaller relevancy.
The last hypothesis is about moves changing the enemy king into check state, making it salient for the attacking player.
Those hypotheses permit to create a dataset way larger than the bottom-up dataset (\etdataset{}). With around two thousand games and an average of fifty round each, we created a bit more than one hundred and twenty thousand new images for training.

Even though this new dataset is substantially big and rich enough, some biases are present.
The first one is that even with the large amount of created configurations, the vast majority of them were openings.
Variety of these openings being limited, there are some redundancies.
The second bias is about the subjectivity of the generated saliency maps.
Our choices about salient areas were driven  by our knowledge of the chess game, observations from our data and examples in the literature \cite{Charness2001,Perceptioninchess,doi:10.1167/17.3.4}.
The last bias concerns the type of moved pieces.
Pawns being the vast majority of them, they are often way more salient because they are  moved more often. Queen, rooks, knights and bishops follow pawns in the salient order. The king being not very often moved nor the center of attention as it is not put in check that often, had a limited saliency even for configurations where it was in danger. Also, because of their key role in the game strategy, players do not need to check the kings' position as often as other pieces (bishops, pawns etc..).

%% file: 05_Results.tex
\section{Experiments and results} \label{chapter:results}

This section presents results obtained using different training strategies.
We are unaware of other research on  visual attention prediction for chess (or other board games) players in the machine learning literature.  
Thus, it is not possible to provide comparative experimental performance evaluation. 
The results presented in this section can provide a baseline for future work on visual attention prediction in similar contexts.

\newcommand{\VI}{V1}
\newcommand{\VII}{V2}
\newcommand{\VIII}{V3}
\newcommand{\VIV}{V4}
\newcommand{\VV}{V5}

\subsection{Quantitative results}
\subsubsection{Evaluation conditions}

To conduct the quantitative evaluation of our approach, we trained several variations of our system. 
The neural network architecture remains the same (see section \ref{chapter:approach}), but the training conditions vary in terms of pretraining and fine training corpora.
At our disposal, we have our 3 datasets \etdataset{}, \augetdataset{} and \gamesdataset{} (see section \ref{chapter:learning}).
We adjoin the CAT2000 dataset~\cite{DBLP:journals/corr/BorjiI15}.
This dataset contains natural images with their saliency maps computed and averaged from eye movements of 120 people.
The underlying idea of using it is to extract general saliency deep features before fine-tuning them on our specific corpora.
The different versions of our training strategies are depicted in table \ref{fig:systems}.

\begin{table}
    \centering
\begin{tabular}{|c|c|c|c|c|}
\cmidrule{2-5}    \multicolumn{1}{r|}{} & \multicolumn{2}{c|}{Pre-training} & \multicolumn{2}{c|}{(Fine-)training} \\
    \midrule
    System & CAT2000 & \gamesdataset{}  & \etdataset{} & \augetdataset{} \\
    \midrule
    \VI{}    &       &  \Checkmark    & \Checkmark    &  \\
    \midrule
    \VII{}    & \Checkmark     &       & \Checkmark     &  \\
    \midrule
    \VIII{}    &       & \Checkmark     &       & \Checkmark \\
    \midrule
    \VIV{}    & \Checkmark     &       &       & \Checkmark \\
    \midrule
    \VV{}    &      &       &       & \Checkmark \\
    \bottomrule
    \end{tabular}
\caption{Training conditions of our system.}
\label{fig:systems}
\end{table}

The first 2 systems are trained on the \etdataset{} dataset. The first one (\VI{}) is pretrained on \gamesdataset{} and the second one (\VII{}) on CAT2000.
The next 2 systems (\VIII{} and \VIV{}) are pretrained the same way but are fine-tuned using the augmented \etdataset{} dataset.
As a reference, the last system is trained on \etdataset{} without pretraining.
All systems are trained during 2000 epochs with a learning rate of 0.01.
\added{
This learning rate has been determined experimentally during the development phase.
It showed the best performance, even compared with a cyclical learning rate approach} \protect\cite{smith2017cyclical}.

\added{
Concerning the usual training/validation/test sets paradigm, the size of the }\protect\etdataset{} \added{dataset restricts the available options.
First, we cannot create a validation set without drastically reducing the size of the training set. 
As we do not want to train and to test on similar saliency maps (i.e. saliency maps on the same chess problem from chess players with about the same expertise), we split training and testing sets using tasks rather than using players.
We kept 9 configurations as training data and used the 2 lasts, the more complex and difficult ones, as test configurations to assess the performance of our proposal.}
This splitting strategy ensures that the neural networks are not trained on the test data.

\subsubsection{Evaluation metrics}\label{sec:metrics}

To measure the quality of the generated saliency maps, we consider them as probability distributions.
Information about available metrics can be found in Bylinskii and al. \cite{metricsreview} and Le Meur and Baccino \cite{LeMeur2013} papers, where the authors explain the interpretation that can be done from those metrics.
In this section we will present the metrics used in this article.

    \textit{Linear Correlation Coefficient (LCC)}. The linear correlation is a statistical method to determine how much two probability distributions are related one to the other. It is computed as the ratio of the covariance of the two distributions and the product of their standard deviations. The value given by the linear correlation ranges between -1 and +1, the closer to +1 or -1 it gets, the more related the two distributions are. Because of its symmetric computation, linear correlation does not make the difference between false positives and false negatives.

    \textit{Similarity (SIM)}. The similarity, also referred as histogram intersection, is the sum of minimums of two normalized saliency maps. A null value indicates that there is no overlap while 1 means that the saliency maps are the same.
    
    \textit{Normalized Scanpath Saliency (NSS)}. NSS was introduced by \citet{Peters2005ComponentsOB} as a correspondence measure between a normalized saliency map and a ground truth at fixation points. It is sensitive to false positives, relative differences in  saliency  across  the  image  and  general   monotonic  transformations. A mean subtraction step is added to gain robustness against any kind of linear transformations. With a value above 0, this metric expresses a correspondence above chance between the two saliency maps, below 0 an anti-correspondence.

    The last metric employed in this paper is the \textit{Area Under the Curve (AUC)}. In visual attention modeling, the main goal is to predict if a pixel is going to be stared at or not, which is basically a binary classification. The ROC (Receiver Operating Characteristic) is in signal detection theory a measure of the trade-off between true and false positives while varying a threshold \cite{FAWCETT2006861}. For saliency evaluation, we consider the saliency map as a binary classifier, varying the threshold to transform probabilities to \{0,1\} values.
    The AUC metric is then the area under the ROC curve.
    Some variants exist in literature: \textit{AUC-{Judd}} \cite{judd,Judd_2012,metricsreview} and \textit{AUC-Borji} \cite{6751224}.
    The difference resides in the way the \textit{AUC-Borji} metric computes the false positive rate as the ratio of values above a threshold sampled at random pixels (as many as fixations, sampled uniformly over the whole image) divided by the number of saliency map pixels at the same threshold.

\subsubsection{Performance assessment}

\begin{table}
    \centering

\begin{tabular}{|c|c|c|c|c|c|}
  \hline
  \makecell{Model/\\Metrics} & \makecell{LCC\\(std)} & \makecell{SIM\\(std)} & \makecell{NSS\\(std)} & \makecell{AUC-Judd\\(std)} & \makecell{AUC-Borji\\(std)} \\
  \hline
  \VI{} & 0.60$\pm$0.10 & 0.68$\pm$0.05 & 0.21$\pm$0.11 & 0.57$\pm$0.07 & 0.67$\pm$0.05 \\
  \VII{} & 0.66$\pm$0.06 & 0.70$\pm$0.03 & 0.23$\pm$0.13 & 0.61$\pm$0.13 & 0.70$\pm$0.08 \\ %
  \VIII{} & 0.64$\pm$0.08 & 0.69$\pm$0.04 & 0.21$\pm$0.10 & 0.62$\pm$0.13 & 0.71$\pm$0.09 \\ %
  \VIV{} & 0.57$\pm$0.10 & 0.65$\pm$0.08 &0.24$\pm$0.09 & 0.69$\pm$0.09 & 0.74$\pm$0.10 \\ %
  \VV{} & 0.54$\pm$0.03 & 0.64$\pm$0.02 & 0.20$\pm$0.10 & 0.66$\pm$0.09 &0.74$\pm$0.06 \\ %
  \hline
\end{tabular}%
\caption{Performances of the different flavors of our neural network, using a 6-fold cross validation on the 11 configurations.} 
 \label{tab:metrics}
\end{table}

The quantitative results of the different training strategies of the proposed neural architecture are presented in table \ref{tab:metrics}.
Looking at this table, one can first draw some general comments.
All systems are better than chance ($NSS>0$) and similarity between the saliency maps and the ground-truth is always over 0.54.
As the correlation ranges from 0.54 (\VV{}, only trained on the non augmented \etdataset{} corpus) to 0.66 (\VII{}), the LCC metric confirms that our approach is promising. 

As a second evaluation point, one can compare the effectiveness of the different (pre)training strategies.
Without pretraining and augmentation, \VV{} is the worst model except for AUC metrics. %
\VII{} and \VIV{} are the best models looking globally at all metrics.
Nevertheless, \VIV{}, pretrained with CAT2000 and then \augetdataset{}, presents better results on AUC metrics, meaning that it has a better discrimination power between fixated/unfixated pixels.
As expected, pretraining our model on the CAT2000 dataset appears to give better results (\VIV{}) even without data augmentation (\VII{}).
Concerning correlation metrics (LCC, SIM, NSS), \VIV{} is not as good as without augmentation (\VII{}) but offers the better performances concerning AUC metrics.

These results show the effectiveness of our investigation.
Systems pretrained with CAT2000 have better general scores.
This important result validates that deep features computed on natural image saliency maps are interesting to process synthetic images like our chess board. 

\subsection{Qualitative results}

The final evaluation consists in qualitative results, i.e. human evaluation of generated saliency maps.
This evaluation will first focus on actual chess problems, then on pure synthetic examples, comparing the best systems (\VIII{} and \VIV{}) of each training strategy in the qualitative evaluation.

\subsubsection{Comparative results}

\begin{figure}[ht!]
    \centering
    \begin{tabular}{@{}c@{\hspace{0.1cm}}c@{\hspace{0.1cm}}}
            {\small Configuration 1 } & {\small Configuration 2}\\
        \includegraphics[width=0.25\linewidth]{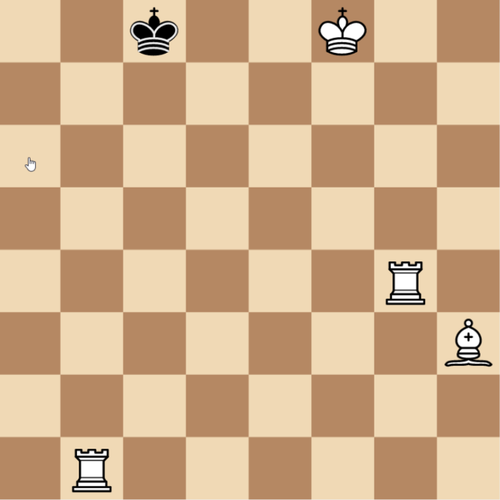}& 
        \includegraphics[width=0.25\linewidth]{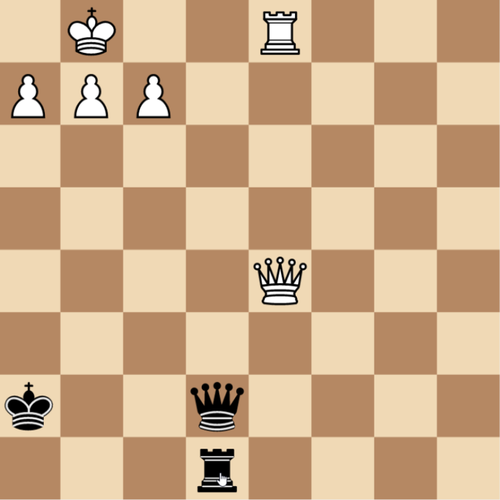}\\
        \includegraphics[width=0.25\linewidth]{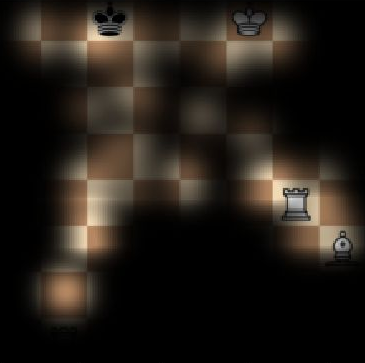}& 
        \includegraphics[width=0.25\linewidth]{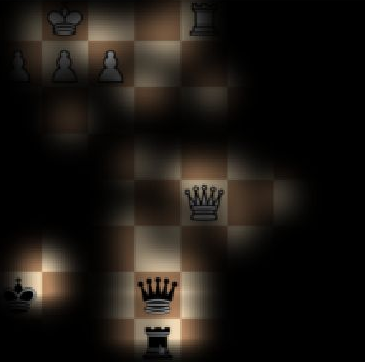}\\
        \includegraphics[width=0.25\linewidth]{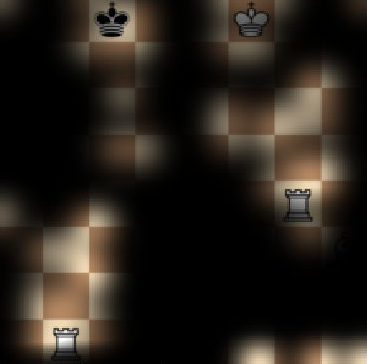}& %
        \includegraphics[width=0.25\linewidth]{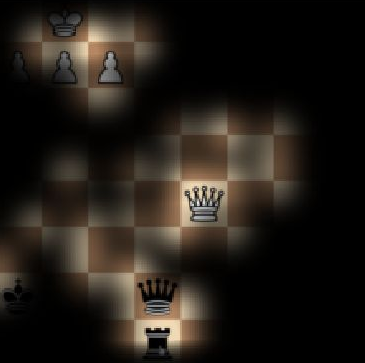}\\        \includegraphics[width=0.25\linewidth]{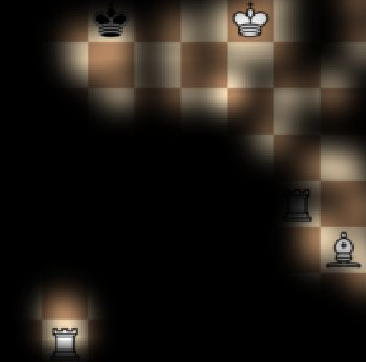}& 
        \includegraphics[width=0.25\linewidth]{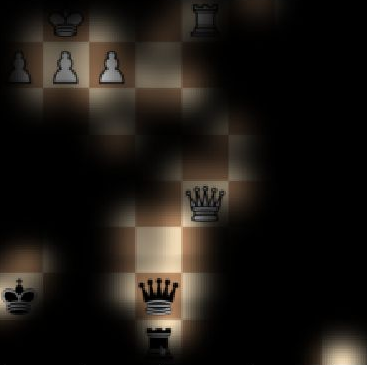}\\
    \end{tabular}
    \caption{Saliency map generation of two models \protect\VIII{} and \VIV{} on two testing configurations. The first row is the chess configuration, the second row the ground-truth from the users' eye tracking data. The third row contains the predictions from \VIII{} pretrained on \gamesdataset{}, the last row from \VIV{} pretrained on CAT2000.}
    \label{fig:modelcomp}
\end{figure}

Saliency map prediction from \VIII{} and \VIV{} models given two unknown chess configurations are depicted on figure \ref{fig:modelcomp} (first row).
On the second row, the figure displays ground-truth results computed from eye tracking data of chess players for these configurations.
Third and fourth rows present respectively saliency map predictions for systems \VIII{} and \VIV{}.
After analysis, several differences can be observed between the two training strategies and the ground-truth.
On one hand, for configuration 1, the network trained on CAT2000 correctly focuses on both kings, rooks and the bishop as actual chess players did.
On the other hand, the other network does not show as much interest for the rook and bishop on the right side of the board.
For the second configuration, both networks  correctly focus on the white king, the rook and the white queen. The black queen and black rook which are the keys to the check mate are salient also for both networks.

\subsubsection{Prediction on synthetic examples}\label{section:examples}

\begin{figure}
\begin{tabularx}{\linewidth}{@{}Y@{}}

\begin{tabular}{@{}c@{\hspace{0.1cm}}c@{\hspace{0.1cm}}c@{}}
{\small} & {\small \VIII{} saliency map} &  {\small \VIV{} saliency map} \\
\includegraphics[width=0.25\linewidth]{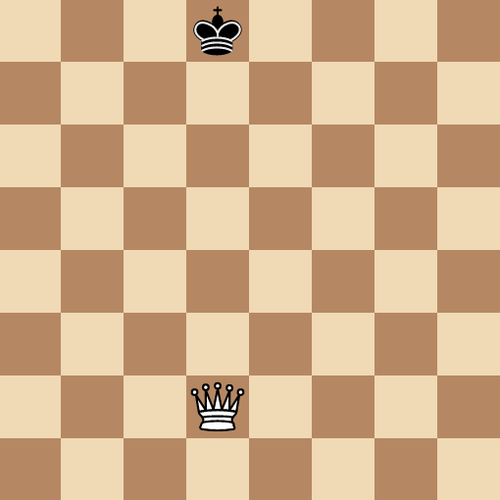} &
\includegraphics[width=0.25\linewidth]{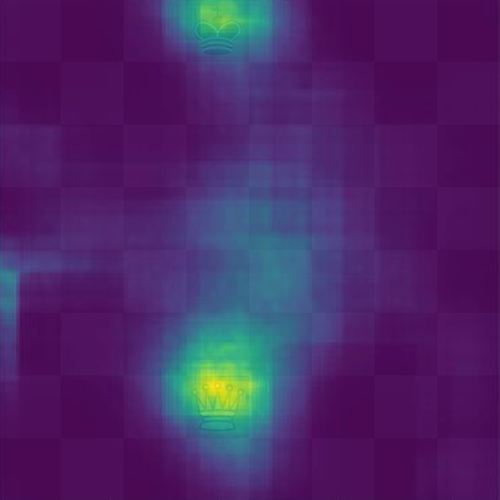} &
\includegraphics[width=0.25\linewidth]{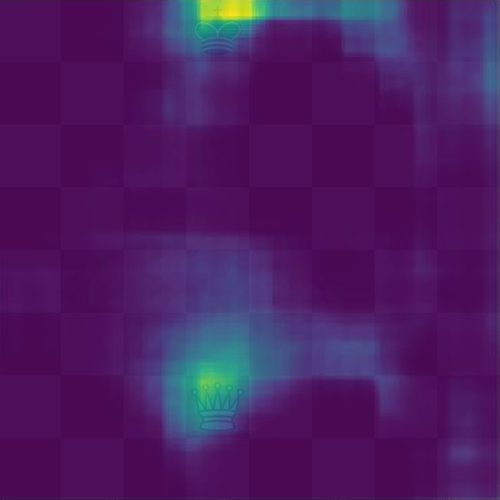} 
\end{tabular}
\caption{An other simple configuration with only one king and its opponent queen}
\label{fig:queenking}
\\ 
\begin{tabular}{@{}c@{\hspace{0.1cm}}c@{\hspace{0.1cm}}c@{}}
{\small} & {\small \VIII{} saliency map} &  {\small \VIV{} saliency map} \\
\includegraphics[width=0.25\linewidth]{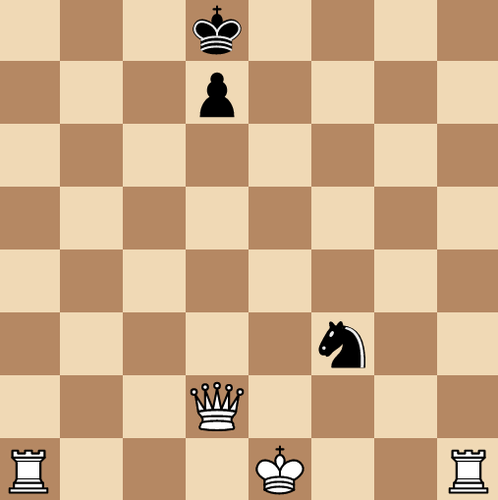}& 
\includegraphics[width=0.25\linewidth]{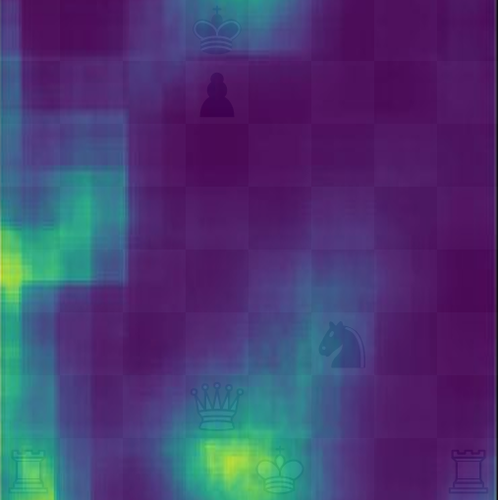}&
\includegraphics[width=0.25\linewidth]{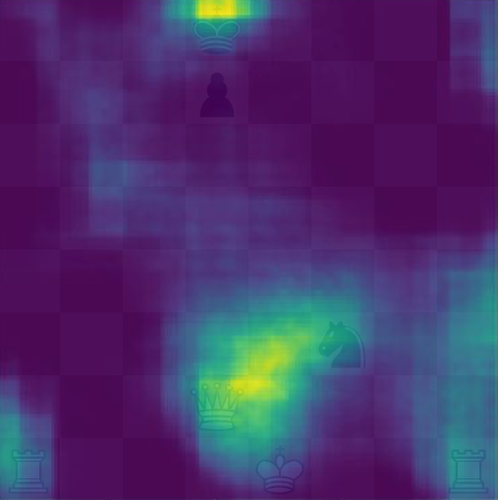}
\end{tabular}
\caption{A bit more complicated configuration. The white king is in check}
\label{fig:withqueen}
\\
\begin{tabular}{@{}c@{\hspace{0.1cm}}c@{\hspace{0.1cm}}c@{}}
{\small} & {\small \VIII{} saliency map} &  {\small \VIV{} saliency map} \\
\includegraphics[width=0.25\linewidth]{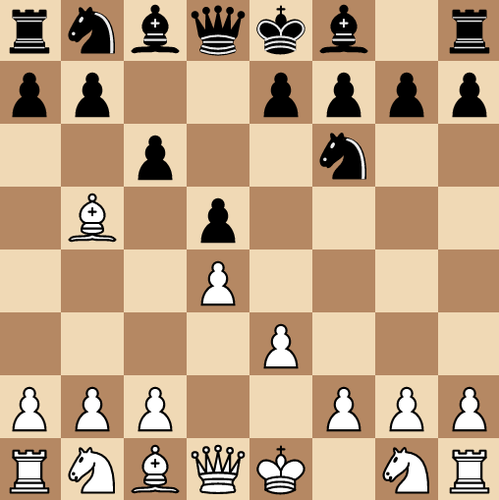}& 
\includegraphics[width=0.25\linewidth]{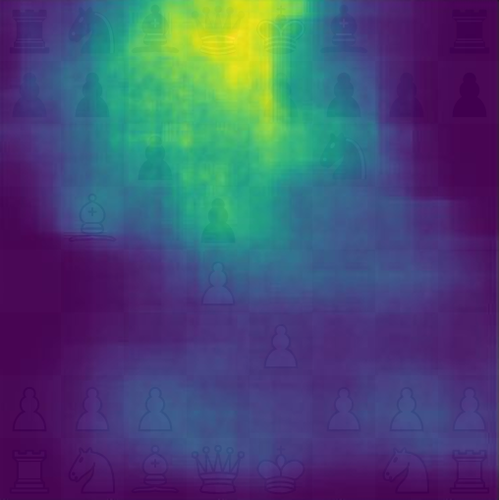}&
\includegraphics[width=0.25\linewidth]{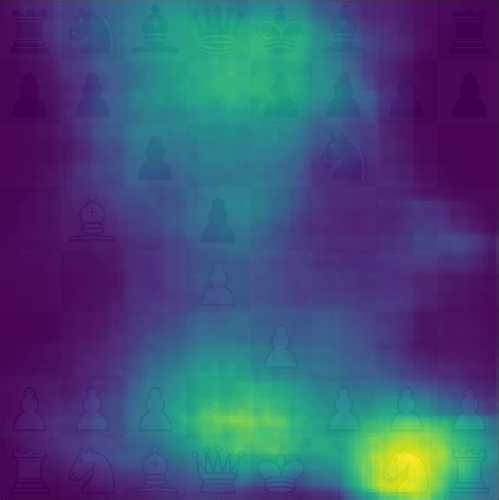}
\end{tabular}
\caption{An opening with a bishop in danger}
\label{fig:opening}
\\
\begin{tabular}{@{}c@{\hspace{0.1cm}}c@{\hspace{0.1cm}}c@{}}
{\small} & {\small \VIII{} saliency map} &  {\small \VIV{} saliency map} \\
\includegraphics[width=0.25\linewidth]{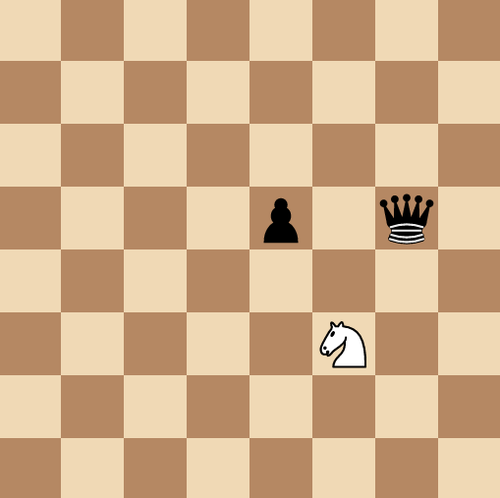}& 
\includegraphics[width=0.25\linewidth]{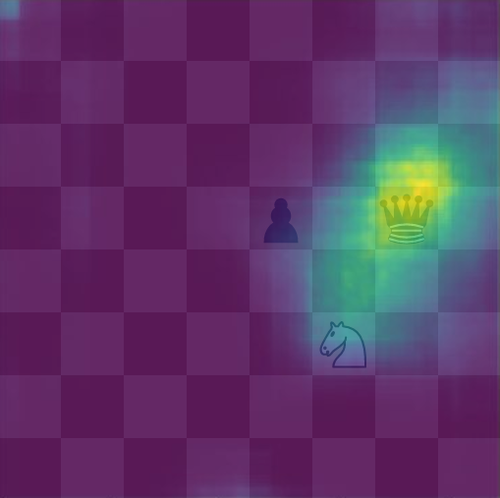}&
\includegraphics[width=0.25\linewidth]{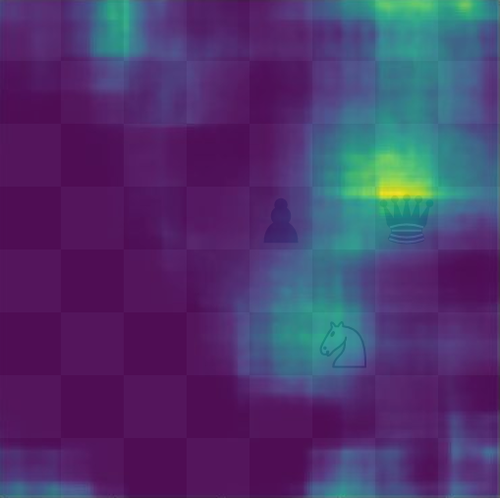}
\end{tabular}
\caption{A choice for a knight}
\label{fig:choice}
\end{tabularx}
\end{figure}

In this section, we will qualitatively assess performances of our networks on several unseen synthetic configurations.
Configurations from figures \ref{fig:queenking} to \ref{fig:choice} were created using the board editor from the Lichess platform.

The figure \ref{fig:queenking} depicts a simple configuration. Here the black king is put in check by the white queen. We would expect fixations on the two pieces and on the vertical line between them showing their relation. \VIII{} is close to this expected result but with some salient areas on the left part of the board. \VIV{} shows less saliency between the two pieces but some on the right part of the board.
A more complicated configurations result is drawn on Figure \ref{fig:withqueen}, showing the white king in check.
The only possibility is to move the king. The expected salient areas would be the white rooks, white king and  white queen and the black knight if we consider the king in check.
If we do not know the check rules, the white queen diagonal and rook's vertical lines would be salient as they can attack the enemy king this way.
Here \VIII{} considers all pieces and the position from where the queen would put the enemy king in check. But rook's verticals are not considered. \VIV{} is considering moving the right rook upward and  the black knight putting the king in check with more saliency.
Figure \ref{fig:opening} is different from previous examples considering the number of pieces. The white bishop in B5 is in danger because of the black pawn in C6. The logic way to solve this problem is to move the bishop either back or to the left. \VIII{}  considers the area surrounding the king/queen of the enemy and including the white bishop (the king was in check the round before). \VIV{} considers the bottom right region (castling area) to be more salient but do not include the white bishop.
The last configuration (Figure \ref{fig:choice}) shows a white knight having two choices. It can either take the pawn and be taken by the queen or take the queen and take the pawn next round.
We would expect the area around the three pieces to be salient as well as the space between them.
\VIII{} saliency map corresponds to these assumptions choosing the queen to be more salient with the knight.
\VIV{} chooses also the queen and the knight but is less focused. Unexpectedly, it adds noisy salient areas everywhere on the board.

These qualitative results offer a glance at what deep neural networks could bring to visual attention prediction for chess players.
As in quantitative results (see previous section), training with the CAT2000 or on synthetic \gamesdataset{} dataset provides different results.
This suggests again that deep features learned on natural images are relevant in our saliency computation for chess board images.
It also highlights that the generated \gamesdataset{} corpus is of interest to learn relationships between pieces.

%% file: 06_Discussion.tex
\section{Discussion} \label{chapter:discussion}

As seen in previous section, the different flavors of our neural network show interesting results and generalizations capability on unseen configurations.
For specific configurations, one can see that pieces and their relationship influence the model prediction as it would do for a chess player, sustained by training on the \augetdataset{} and \gamesdataset{} corpora.
Using a well-known visual attention prediction dataset (CAT2000) \cite{DBLP:journals/corr/BorjiI15} to train the decoder part to transform features to salient regions, led to better performances on standard metrics and results closer to the ground-truth.
However, it was a bit struggling to generalize to new configurations, displaying some salient areas around the boundaries of the board.
Nonetheless, pretraining the decoder part of the network with such dataset seems to be a good way to initialize the network weights.
This may be explained by the fact that the network encoder is initialized using VGG weights trained on the ImageNet dataset \cite{DBLP:journals/corr/SimonyanZ14a}.

The reported results may hide that there are still rooms for improvement, the underlying bottleneck of these researches on chess remaining the amount of available data.
The available recordings of eye tracking data in chess context \cite{thomasguntz} represent a huge amount of work and hours of data.
Nevertheless, they contain few configurations and around 30 players recorded in actual chess problem solving tasks.
This is not enough for generalization and performance, moreover when using deep learning techniques.
This explains the benefit from pre-training on the CAT2000 corpus and from our generated game dataset (\gamesdataset{}).

The first way to improve current results would be to record eye tracking data for players playing full games during online tournaments or even when casually training for instance.
With such data, the training corpus will contain more configurations and get more eye-tracking data and from a larger panel of players' expertise.
With more data, it will be possible to create different deep learning networks based on the players' skill.
It has been shown in previous work that for board games such as chess, players do not look at the game the same way depending on their expertise \cite{Charness2001,Perceptioninchess,doi:10.1167/17.3.4}. Doing this separation, one could have different networks pretrained on the whole data, but adjusted on data from different player groups (beginner, intermediates, experts for instance). The resulting neural networks would be able to model visual attention for their respective expertise group.

Using the \gamesdataset{} dataset showed to be efficient for producing saliency map when no visual attention prediction dataset is available.
Previous depicted results let us assess some points about it:
\begin{itemize}[leftmargin=*]
    \vspace{-0.5em}\item There is a content bias. This \gamesdataset{} corpus contains more openings, so there are more salient areas on the player side. This could however  be partially overthrown after training on the augmented dataset (\augetdataset).
    \item The results from the network pretrained on this dataset are showing smaller salient areas focused on pieces. This is a direct consequence of how the dataset has been created, but is interesting as it corresponds more to how beginner/intermediate players would look at the board. %
    \item The network seems to recognize more pieces and their relations when pretrained using this dataset.
\end{itemize}

The \gamesdataset{} corpus can benefit from the immense number of chess games available online.
A second way to improve these researches would be to enhance the generation process of this dataset (depicted in section \ref{chapter:learning}).
The game dataset generation can be improved by defining a strategy to select interesting parts of games to produce a more balanced corpus, maybe integrating
players' statistics to produce saliency maps reflecting what is known about the link between fixations and expertise.

%% file: 07_Conclusion.tex
\vspace{-0.8em}
\section{Conclusion} \label{chapter:conclusion}

Training a CNN for visual attention prediction is deeply task dependent and thus is not a solved problem notably in chess context.
The promising quantitative and qualitative result reported here show that using data augmentation and task driven data, visual attention prediction of chess players is possible.
The proposed network architecture with an auto-encoder is able to learn and to predict saliency map in the chess context.
When focusing on a Bottom-up approach using raw images as input, the model was able to extract features and patterns from limited data.
Data augmentation was found to improve performances when pretraining on the synthetic \gamesdataset{} dataset, enhancing locations of salient areas rather than extracting correspondence between the saliency maps and the ground-truth, learning smaller and more general patterns. 
\added{After gathering more chess games to cope with the lack of eye-tracking data, we should be able to use the proposed neural network architecture as input for researches on analyzing and interpreting multimodal signals of players engaged in solving challenging problems.}

In this study, we have opened the investigation concerning visual attention prediction for chess players.
It may be possible to apply similar approaches to other board games such as GO, where the model used and its training would be the same, with just a variation of the data and game rules.
The proposed visual attention model showcases good performances even if there are still rooms for improvement, as working on quality and quantity of our training corpus.
The novelty of our work makes it difficult to compare to existing techniques, but current results and ideas could be used as a baseline for future work.

%% file: 08_ACKS.tex
\section{Acknowledgments}

This research has been funded by the French ANR project CEEGE (ANR-15-CE23-0005), and was made possible by the use of equipment provided by ANR Equipment for Excellence Amiqual4Home (ANR-11-EQPX-0002).